\documentclass[10pt]{article}

\usepackage[utf8]{inputenc}
\usepackage[T1]{fontenc}
\usepackage{graphicx}
\usepackage{amsmath,amssymb}
\usepackage{booktabs}
\usepackage{longtable}
\usepackage{hyperref}
\usepackage{multirow}
\usepackage{caption}
\usepackage{adjustbox}
\usepackage{array}
\usepackage{placeins}
\usepackage{float}
\usepackage{subcaption}
\usepackage{tikz}
\usetikzlibrary{shapes, arrows.meta, positioning}

\title{Digital Image Forgery Detection Using Transfer Learning}

\author{
Fatma Betul BUYUK$^{*1}$ \and
Gozde KARATAS BAYDOGMUS$^{2,3}$ \and
Ali BULDU$^{1}$ \and
Ayaulym TULENDIYEVA$^{1}$ \and
Zhuldyz BAIZHUMANOVA$^{1}$ \\
[6pt]
\small $^{1}$Marmara University, Department of Computer Engineering,
\small Istanbul, TURKEY \\
\small $^{2}$Loyola University Chicago, Department of Computer Science, Chicago, USA \\
\small $^{3}$Biruni University, Department of Computer Engineering,
\small Istanbul, TURKEY \\
}

\date{}

\begin{document}

\maketitle

\begin{abstract}
The increasing availability of advanced image editing tools has led to a significant rise in manipulated digital content, posing serious challenges for digital forensics and information security. This study presents a transfer learning–based framework for digital image forgery detection that integrates compression-aware feature enhancement with deep convolutional neural network (CNN) architectures.

The proposed approach introduces a hybrid input representation that combines RGB images with compression difference-based features (FDIFF), explicitly highlighting subtle manipulation artifacts that are often difficult to detect. In addition, a model-specific adaptive threshold optimization strategy based on the Youden Index is employed to improve classification reliability by achieving a better balance between true positive and false positive rates.

Experiments conducted on the CASIA v2.0 dataset using multiple pretrained CNN architectures, including DenseNet121, VGG16, ResNet50, EfficientNetB0, MobileNet, and InceptionV3, demonstrate the effectiveness and robustness of the proposed framework. The models are evaluated using comprehensive performance metrics such as accuracy, precision, recall, F1-score, Matthews correlation coefficient (MCC), and area under the ROC curve (AUC).

The results show that DenseNet121 achieves the highest accuracy and AUC, while ResNet50 provides the most balanced and reliable predictions with the highest MCC. The findings emphasize that relying solely on accuracy is insufficient for forensic applications, where minimizing false negatives is critical. Overall, the proposed framework improves the visibility of manipulation artifacts and enhances classification robustness, making it suitable for real-world digital image forgery detection scenarios.
\end{abstract}


\section{Introduction}

The widespread accessibility of advanced digital imaging technologies and image editing tools has significantly increased both the production and dissemination of manipulated visual content across social media, journalism, and forensic domains. This growing prevalence has intensified concerns regarding the authenticity and reliability of visual information. Consequently, digital image forgery detection has become a critical research topic in the fields of digital forensics and information security. Early studies in this area primarily focused on identifying statistical inconsistencies within images~\cite{27}, laying the foundation for subsequent research.

Conventional image forgery detection methods typically rely on handcrafted features designed to capture irregularities in color distribution, texture patterns, or compression artifacts. Although these approaches have demonstrated effectiveness in certain scenarios, they often struggle to detect complex and subtle manipulations, particularly when images are subjected to multiple post-processing operations such as compression, scaling, and rotation~\cite{3,4}. These limitations have motivated the transition toward deep learning–based approaches, which enable automatic extraction of discriminative features from large-scale datasets. 

\begin{figure}[H]
\centering
\includegraphics[width=0.85\textwidth]{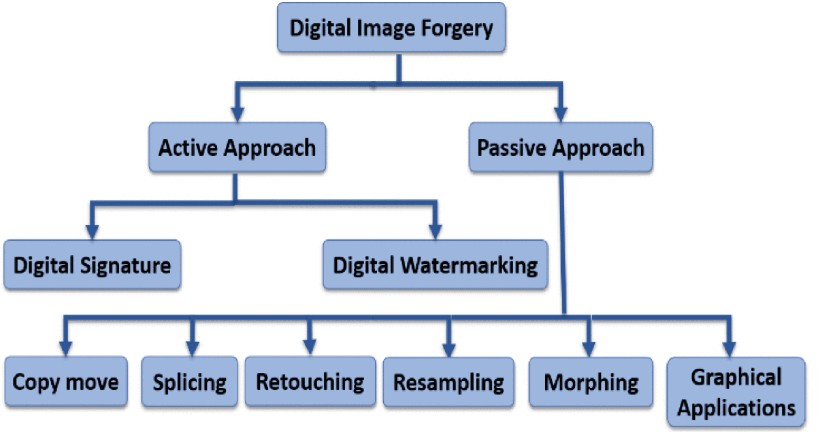}
\caption{Classification of digital image forgery techniques into active and passive approaches.~\cite{16}}
\label{fig:forgery_types}
\end{figure}

Digital image forgery techniques are generally categorized into active and passive approaches. Active methods depend on pre-embedded information such as digital signatures or watermarks, whereas passive methods identify inconsistencies without requiring prior information. Passive approaches encompass various manipulation types, including copy-move, splicing, retouching, resampling, and morphing, as illustrated in Fig.~\ref{fig:forgery_types}.

In recent years, convolutional neural networks (CNNs) have achieved remarkable success in image classification and forgery detection tasks due to their ability to learn hierarchical feature representations. However, training deep CNN models from scratch typically requires large annotated datasets and substantial computational resources. To overcome these challenges, transfer learning has emerged as an effective solution, allowing pretrained models to be adapted to new tasks with limited data while maintaining strong performance~\cite{5,6}. Various studies have successfully applied transfer learning to digital image forgery detection using architectures such as VGG, ResNet, DenseNet, and EfficientNet~\cite{7,8}. 

\begin{figure}[H]
\centering
\includegraphics[width=0.85\textwidth]{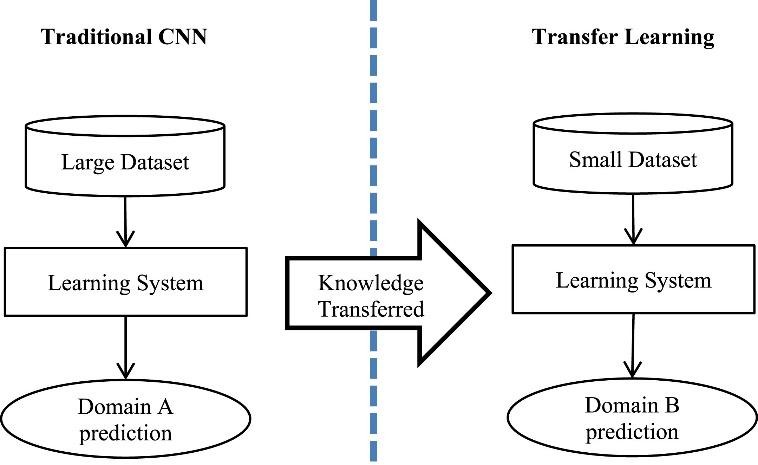}
\caption{Illustration of transfer learning, where knowledge learned from large datasets is transferred to improve performance on smaller datasets~\cite{8}.}
\label{fig:transfer_learning}
\end{figure}

Recent research has also focused on developing deep learning–based feature extraction techniques tailored specifically for manipulation detection tasks~\cite{21,26}. The concept of transfer learning, where knowledge acquired from large-scale datasets is transferred to improve performance on smaller datasets, is illustrated in Fig.~\ref{fig:transfer_learning}.

\begin{figure}[H]
\centering
\includegraphics[width=0.8\textwidth]{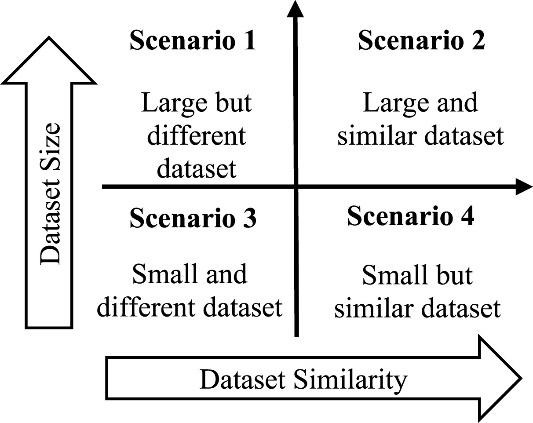}
\caption{Different transfer learning scenarios based on dataset size and similarity~\cite{8}.}
\label{fig:tl_scenarios}
\end{figure}

The effectiveness of transfer learning is influenced by both the size of the dataset and the similarity between source and target domains. As shown in Fig.~\ref{fig:tl_scenarios}, transfer learning yields optimal performance when the datasets are similar, even if the target dataset is relatively small. In contrast, significant domain differences may limit the effectiveness of transferred knowledge.

Despite these advancements, many existing approaches rely directly on raw RGB images as input, which may not sufficiently emphasize subtle manipulation artifacts. In particular, compression inconsistencies introduced during tampering are often not explicitly represented in standard input formats, potentially limiting the ability of deep learning models to detect fine-grained forgery traces.

To address this limitation, this study proposes a transfer learning–based framework that integrates compression-based feature enhancement with deep CNN architectures. Specifically, a difference-based representation (FDIFF) is introduced to highlight compression artifacts by computing the difference between original and recompressed images. This representation enhances the visibility of subtle manipulation traces, enabling CNN models to learn more discriminative and task-relevant features.

Furthermore, unlike conventional approaches that employ fixed classification thresholds, this study adopts an adaptive threshold selection strategy based on the Youden Index. This strategy improves classification reliability by optimizing the trade-off between true positive and false positive rates, which is particularly critical in forensic applications where minimizing false negatives is essential.

The proposed framework is evaluated on the CASIA v2.0 dataset using multiple pretrained CNN architectures, including DenseNet121, VGG16, ResNet50, EfficientNetB0, MobileNet, and InceptionV3. Model performance is assessed using comprehensive evaluation metrics such as accuracy, precision, recall, F1-score, Matthews correlation coefficient (MCC), and area under the ROC curve (AUC).

The main contributions of this study can be summarized as follows:

\begin{itemize}
\item A compression-based feature enhancement approach (FDIFF) is proposed to improve the visibility of manipulation artifacts in forged images.
\item An adaptive threshold optimization strategy based on the Youden Index is introduced to enhance classification reliability.
\item A comprehensive evaluation of multiple pretrained CNN architectures is conducted using both traditional and advanced performance metrics.
\item The study demonstrates that balanced performance metrics such as MCC provide a more realistic evaluation of forgery detection systems compared to accuracy alone.
\end{itemize}

The remainder of this paper is organized as follows. Section II reviews related work on digital image forgery detection. Section III describes the proposed methodology. Section IV presents the experimental results and evaluation. Section V discusses the findings, Section VI concludes the paper, and Section VII outlines future research directions. Finally, the acknowledgment section presents the funding information.

\section{Related Work}

Digital image forgery detection has attracted significant attention in computer vision and digital forensics, with existing approaches broadly categorized into traditional handcrafted feature-based methods and deep learning–based techniques.

Earlier studies primarily relied on handcrafted features to identify inconsistencies in image properties such as color distribution, texture patterns, and compression artifacts. These approaches typically employed statistical analysis and signal processing techniques to reveal traces of manipulation~\cite{3,4}. While these methods are effective for detecting certain types of forgeries, their performance tends to degrade when handling complex manipulations or images subjected to multiple post-processing operations.

The emergence of deep learning has significantly transformed this field, with convolutional neural networks (CNNs) becoming a dominant paradigm for forgery detection. Unlike traditional methods, CNN-based models can automatically learn multi-level feature representations directly from data, enabling more effective detection of subtle manipulation artifacts that are difficult to capture using handcrafted features~\cite{5,6}. Architectures such as VGG, ResNet, DenseNet, and EfficientNet have been widely adopted due to their strong feature representation capabilities.

Recent studies have further improved detection performance by incorporating hybrid deep learning architectures. For instance, Bappy et al.~\cite{12} introduced a CNN-LSTM framework that performs both detection and localization of manipulated regions. Similarly, Elaskily et al.~\cite{3} employed ConvLSTM-based models for copy-move forgery detection, demonstrating improved modeling of spatial and temporal dependencies.

Transfer learning has also emerged as an effective strategy to mitigate the challenges associated with training deep networks from scratch. By utilizing pretrained models on large-scale datasets such as ImageNet, transfer learning facilitates efficient feature extraction while reducing computational requirements~\cite{7,8}. Several studies have reported that transfer learning–based methods achieve competitive performance in image forgery detection tasks~\cite{16,18}.

In addition to classification-based approaches, various deep learning techniques have been proposed to enhance manipulation detection. These include autoencoder-based anomaly detection methods~\cite{19} and feature learning strategies specifically designed to capture manipulation traces~\cite{21}. Furthermore, spatial structure modeling and forgery localization techniques have been explored to improve detection accuracy and interpretability~\cite{22,23,24}.

Despite these advancements, many existing methods still rely directly on raw RGB images as input, which may not sufficiently highlight subtle manipulation artifacts~\cite{2,7}. In particular, compression inconsistencies introduced during tampering are often underrepresented in standard input formats, limiting the effectiveness of CNN models in identifying fine-grained forgery traces~\cite{25}.

Another limitation in the literature is the predominant focus on maximizing overall accuracy, often at the expense of other critical performance metrics such as MCC and the analysis of false negatives~\cite{7,21}. In digital forensic applications, false negatives are particularly critical, as failing to detect manipulated images may lead to significant real-world consequences.

To overcome these limitations, this study proposes a framework that integrates compression-based feature enhancement with transfer learning. The proposed FDIFF representation explicitly emphasizes compression artifacts, enabling improved detection of subtle manipulations~\cite{25}. In addition, an adaptive threshold selection strategy based on the Youden Index is employed to enhance classification reliability and reduce critical errors.

In contrast to many existing studies, this work prioritizes not only accuracy but also balanced performance evaluation and classification reliability, providing a more comprehensive and realistic assessment framework for digital image forgery detection systems~\cite{19,24}. Furthermore, recent developments in the field have also explored lightweight and task-specific architectures, including U-Net–based models~\cite{17} and DeepFake detection approaches~\cite{28}.

\section{Method}

This study proposes a transfer learning–based framework for digital image forgery detection that integrates compression-based feature enhancement with deep convolutional neural network (CNN) architectures. The proposed methodology consists of three main stages: (i) compression-based feature enhancement, (ii) feature extraction using pretrained CNN models, and (iii) classification with adaptive threshold optimization.

\subsection{Transfer Learning}

Transfer learning provides an efficient approach for digital image forgery detection by leveraging knowledge learned from large-scale image datasets. Instead of training deep neural networks from scratch, pretrained models are adapted to the target domain, enabling faster convergence and improved generalization performance~\cite{6,16,18}.

In this study, pretrained CNN architectures including DenseNet121, VGG16, ResNet50, EfficientNetB0, MobileNet, and InceptionV3 are employed. The convolutional layers of these models are retained to preserve learned feature representations, while the final fully connected layers are replaced with task-specific classification layers for binary classification~\cite{7,16}.

Fine-tuning is applied to adapt the pretrained models to the forgery detection task. This process allows the models to learn domain-specific features related to manipulation artifacts while benefiting from the general feature representations learned from large-scale datasets~\cite{6,7}.

\subsection{Compression-Based Feature Enhancement}

A key contribution of this study is the use of compression-based feature enhancement to improve the detection of subtle manipulation artifacts. Digitally manipulated regions often exhibit different compression characteristics compared to authentic regions due to recompression and editing operations~\cite{25}.

To exploit this property, a difference-based representation is constructed using the following components:

\begin{itemize}
\item \textbf{Original Image ($O$):} The unaltered image.
\item \textbf{Forged Image ($F$):} The manipulated image.
\item \textbf{Compressed Forged Image ($F_{comp}$):} A recompressed version of the forged image.
\item \textbf{Difference Image ($F_{diff}$):} Computed as:
\end{itemize}

\begin{equation}
F_{diff} = F - F_{comp}
\end{equation}

The resulting $F_{diff}$ representation enhances compression inconsistencies between manipulated and authentic regions. Unlike raw RGB inputs, this representation explicitly emphasizes subtle manipulation artifacts introduced during tampering, enabling CNN models to learn more discriminative features for forgery detection~\cite{25}.

All images are resized to $224 \times 224$ pixels (and $299 \times 299$ for InceptionV3) to match the input requirements of pretrained models~\cite{9,10,11,14}.

\subsection{Pretrained CNN Architectures}

To evaluate the effectiveness of the proposed approach, multiple pretrained CNN architectures are utilized:

\begin{itemize}
\item DenseNet121~\cite{14}
\item ResNet50~\cite{9}
\item VGG16~\cite{10}
\item EfficientNetB0~\cite{13}
\item MobileNet~\cite{15}
\item InceptionV3~\cite{11}
\end{itemize}

Each model is fine-tuned on the enhanced input representation combining RGB images with the $F_{diff}$ features~\cite{7,16}. The convolutional base is preserved, while the classification head is modified to adapt to the binary classification task~\cite{7}. This ensures that the models can effectively leverage both general and domain-specific feature representations~\cite{6,7}.

\subsection{Classification Architecture}

The classification head added to each pretrained model consists of:

\begin{itemize}
\item Global Average Pooling layer to reduce feature maps
\item A fully connected layer with 512 neurons using ReLU activation
\item Dropout layer with a rate of 0.5 to reduce overfitting
\item A final sigmoid activation layer for binary classification
\end{itemize}

\begin{figure}[H]
\centering
\begin{tikzpicture}[
node distance=1.6cm,
every node/.style={draw, rectangle, rounded corners, align=center, minimum width=3.5cm, minimum height=0.9cm},
arrow/.style={->, thick}
]

\node (input) {Input Image};

\node (fdiff) [below of=input] {FDIFF + RGB\\Combination};

\node (cnn) [below of=fdiff, fill=blue!20] {Pretrained CNN\\(DenseNet, ResNet, etc.)};

\node (gap) [below of=cnn, fill=yellow!30] {Global Average Pooling};

\node (dense) [below of=gap, fill=orange!30] {Dense (512, ReLU)};

\node (dropout) [below of=dense, fill=green!30] {Dropout (0.5)};

\node (output) [below of=dropout, fill=red!30] {Output Layer\\(Sigmoid)};

\draw[arrow] (input) -- (fdiff);
\draw[arrow] (fdiff) -- (cnn);
\draw[arrow] (cnn) -- (gap);
\draw[arrow] (gap) -- (dense);
\draw[arrow] (dense) -- (dropout);
\draw[arrow] (dropout) -- (output);

\end{tikzpicture}
\caption{Proposed transfer learning-based classification architecture incorporating FDIFF-based feature enhancement and pretrained CNN models.}
\label{fig:model_architecture}
\end{figure}
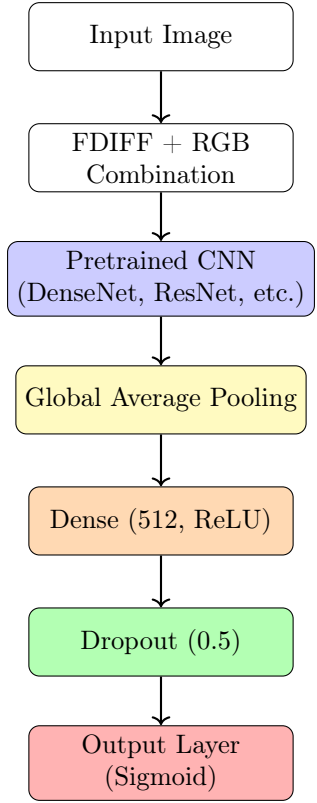

This design follows widely adopted transfer learning practices for adapting pretrained CNN models to classification tasks~\cite{6,7}. The overall architecture of the proposed classification model is illustrated in Fig.~\ref{fig:model_architecture}, highlighting the integration of pretrained convolutional layers and fully connected layers for binary classification.

This architecture enables effective learning of discriminative features while maintaining generalization capability.

\subsection{Dataset}

The CASIA v2.0 dataset, obtained from a publicly available Kaggle repository~\cite{29}, is used in this study. The dataset contains both authentic and manipulated images, with a total of 12,614 images, including 7,491 authentic and 5,123 tampered samples~\cite{2,16}. It includes multiple forgery types such as copy-move and splicing, providing a diverse benchmark for evaluation.

The dataset is randomly divided into training and testing sets using an 80:20 ratio with stratified sampling to preserve class distribution. During training, a portion of the test set is used as validation data to monitor model performance and guide early stopping.

This setup was used solely for monitoring training behavior and does not influence the final evaluation, which is performed on the test data. However, using the test set for validation may introduce a potential bias in performance estimation. Therefore, the reported results should be interpreted with this consideration. Future studies may incorporate a dedicated validation split to eliminate this limitation and further improve evaluation reliability.

\subsection{Evaluation Metrics}

To evaluate model performance, multiple metrics are used:

\begin{itemize}
\item Accuracy
\item Precision
\item Recall
\item F1-score
\item Matthews Correlation Coefficient (MCC)
\item Area Under the Curve (AUC)
\end{itemize}

Unlike many studies that rely solely on accuracy, this work incorporates MCC and AUC to provide a more balanced and reliable evaluation of classification performance, particularly in imbalanced datasets.

\subsection{Adaptive Threshold Optimization}

Instead of using a fixed classification threshold (e.g., 0.5), this study employs an adaptive threshold selection strategy based on the Youden Index~\cite{30}. The optimal threshold is determined by maximizing the difference between the true positive rate (TPR) and false positive rate (FPR), defined as:

\begin{equation}
J = TPR - FPR
\end{equation}

The Youden Index is a widely used statistical measure for evaluating diagnostic tests and binary classifiers, as it provides an effective balance between sensitivity and specificity~\cite{5,30}. It has been successfully applied in various machine learning and medical decision-making problems to determine optimal decision thresholds~\cite{5}.

Unlike conventional approaches that apply a fixed global threshold across all models~\cite{7,21}, the proposed framework determines a model-specific optimal threshold for each CNN architecture individually. As presented in Table~\ref{tab:Evaluation Metrics with Model-Specific Optimal Thresholds}, each model yields a different optimal threshold due to variations in their output probability distributions.

The optimal thresholds obtained for DenseNet121, VGG16, ResNet50, EfficientNetB0, MobileNet, and InceptionV3 are 0.395, 0.426, 0.338, 0.461, 0.419, and 0.361, respectively. These variations indicate that each model exhibits distinct decision characteristics and confidence calibration, which has also been observed in prior deep learning-based classification studies~\cite{7,21}.

For each trained model, the ROC curve is computed on the test set, and the threshold corresponding to the maximum Youden Index is selected~\cite{30}. This model-specific threshold is then used to convert predicted probabilities into binary class labels.

This adaptive and model-specific thresholding strategy improves classification reliability by accounting for differences in score distributions across architectures~\cite{5}. It enhances the balance between sensitivity and specificity and significantly reduces false negatives, which is particularly critical in digital forensic applications~\cite{2,19}. Consequently, the proposed approach increases the robustness and practical applicability of the forgery detection framework.

These findings further demonstrate that threshold calibration is a critical component in deep learning-based forensic systems, as different architectures produce varying confidence distributions~\cite{5,7}.

\section{Experimental Results and Evaluation}

\subsection{Evaluation of Model Training and Testing}

To evaluate the effectiveness of the proposed model, a comprehensive analysis was conducted using performance metrics, the Receiver Operating Characteristic (ROC) curve, and confusion matrices. These tools are widely used to assess classification performance in binary decision problems and provide detailed insights into model reliability and discriminative capability~\cite{1,3}. The evaluation employed key metrics, including Test Accuracy, Precision, Recall, F1-Score, Matthews Correlation Coefficient (MCC), and Area Under the Curve (AUC), which together offer a comprehensive understanding of classification performance.

The model configuration is defined as follows. The input image size was set to 224$\times$224 pixels (and 299$\times$299 for InceptionV3), and the initial weights were pre-trained on the ImageNet dataset~\cite{9,10}. The model was trained for a maximum of 100 epochs, incorporating an early stopping mechanism that monitored the validation loss with a patience of 10 epochs~\cite{12}. The Adam optimizer was employed with a learning rate of $1 \times 10^{-5}$, and binary cross-entropy was used as the loss function~\cite{7}. The dataset was divided into training and testing sets using an 80:20 ratio. In addition, the test set was also used as validation data during training to monitor model performance and prevent overfitting~\cite{5,16}. The final evaluation results are presented in Table~\ref{tab:Evaluation Metrics with Model-Specific Optimal Thresholds}.

In addition to conventional performance metrics, the results indicate that each model operates with a distinct optimal threshold, as shown in Table~\ref{tab:Evaluation Metrics with Model-Specific Optimal Thresholds}. The variation in threshold values (ranging from 0.338 to 0.461) indicates that different CNN architectures produce different probability distributions and decision boundaries, a phenomenon commonly observed in deep learning-based classification systems~\cite{7,21}.

Notably, models such as ResNet50 achieve optimal performance with lower threshold values, suggesting a more conservative decision behavior that favors reducing false negatives. In contrast, models with higher threshold values, such as EfficientNetB0, tend to require stronger confidence for positive classification. Such variations in decision behavior are consistent with findings reported in prior studies highlighting differences in model calibration and confidence estimation across architectures~\cite{5,7}.

These findings demonstrate that applying a fixed threshold (e.g., 0.5) would lead to suboptimal performance across models~\cite{7,21}. Therefore, the use of model-specific adaptive thresholds plays a critical role in improving classification balance and overall reliability, particularly in sensitive domains such as digital forensics where minimizing false negatives is essential~\cite{2,19}.

\begin{table*}[h]
\centering
\caption{Performance Evaluation of CNN Models with Model-Specific Optimal Thresholds}
\label{tab:Evaluation Metrics with Model-Specific Optimal Thresholds}
\resizebox{\textwidth}{!}{%
\begin{tabular}{lcccccc}
\toprule
\textbf{Metric} & \textbf{DenseNet121} & \textbf{VGG16} & \textbf{ResNet50} & \textbf{EfficientNetB0} & \textbf{MobileNet} & \textbf{InceptionV3} \\
\midrule
Test Accuracy & 78.4\% & 70.9\% & 77.6\% & 75.1\% & 74.9\% & 74.5\% \\
Precision     & 77.3\% & 69.3\% & 77.1\% & 73.7\% & 74.1\% & 73.9\% \\
Recall        & 78.4\% & 70.9\% & 77.6\% & 75.1\% & 74.9\% & 74.5\% \\
F1-Score      & 77.0\% & 68.5\% & 77.2\% & 72.9\% & 73.2\% & 73.6\% \\
MCC           & 0.593 & 0.434 & 0.598 & 0.517 & 0.520 & 0.525 \\
AUC           & 0.841 & 0.779 & 0.827 & 0.811 & 0.806 & 0.820 \\
Threshold     & 0.395 & 0.426 & 0.338 & 0.461 & 0.419 & 0.361 \\
Input Size    & 224$\times$224$\times$3 & 224$\times$224$\times$3 & 224$\times$224$\times$3 & 224$\times$224$\times$3 & 224$\times$224$\times$3 & 299$\times$299$\times$3 \\
\bottomrule
\end{tabular}}
\end{table*}

To comprehensively evaluate the effectiveness of various deep learning architectures, six pre-trained models were compared using the metrics presented in Table~\ref{tab:Evaluation Metrics with Model-Specific Optimal Thresholds}.

\begin{itemize}

\item \textbf{Test Accuracy:} DenseNet121 achieved the highest accuracy (78.4\%), closely followed by ResNet50 (77.6\%), indicating stronger overall classification capability.

\item \textbf{Precision:} Precision values ranged from 69.3\% (VGG16) to 77.3\% (DenseNet121). Higher precision values indicate improved ability to reduce false positives.

\item \textbf{Recall:} DenseNet121 and ResNet50 achieved the highest recall values, demonstrating better capability in detecting forged images.

\item \textbf{F1-Score:} ResNet50 achieved the highest F1-score (77.2\%), indicating balanced performance.

\item \textbf{MCC:} ResNet50 achieved the highest MCC value (0.598), reflecting the most reliable and balanced classification performance~\cite{3}.

\item \textbf{AUC:} DenseNet121 achieved the highest AUC (0.841), indicating strong discriminative capability across thresholds~\cite{5}.

\item \textbf{Threshold:} The optimal threshold values vary across models, ranging from 0.338 (ResNet50) to 0.461 (EfficientNetB0). These differences indicate that each architecture produces distinct probability distributions and decision boundaries, which has been widely observed in deep learning-based classification systems~\cite{7,21}. Models with lower threshold values, such as ResNet50, tend to favor sensitivity and reduce false negatives, while models with higher thresholds require stronger confidence for positive classification. This result highlights that using a fixed threshold (e.g., 0.5) would lead to suboptimal performance and confirms the importance of model-specific adaptive thresholding~\cite{5,30}.

\item \textbf{Input Size:} The larger input size of InceptionV3 did not result in superior performance.

\end{itemize}

\subsubsection{Model Comparison:}

Unlike many previous studies, InceptionV3 did not achieve the best performance in this study. Instead, DenseNet121 and ResNet50 consistently outperformed other architectures across multiple evaluation metrics. In particular, ResNet50 achieved the highest MCC value, indicating more reliable classification performance, especially in reducing critical classification errors.

In digital image forensics, minimizing false negatives is of primary importance, as undetected manipulated images may lead to misinformation and security risks~\cite{2}. In this context, ResNet50 demonstrates superior reliability despite not having the highest accuracy.

\textbf{Performance Interpretation:}

Although the achieved accuracy values are lower than some existing studies, this does not necessarily indicate weaker performance. Many previous works primarily focus on maximizing accuracy, often overlooking other critical performance aspects~\cite{7,21}.

In contrast, this study evaluates model performance using multiple complementary metrics, including MCC and AUC, to provide a more comprehensive and realistic assessment~\cite{5}. In forensic applications, false negatives represent critical errors; therefore, models with more balanced performance are more suitable for real-world deployment~\cite{2,19}.

Furthermore, the use of compression-based feature enhancement and adaptive threshold optimization introduces a more challenging and realistic evaluation setting compared to conventional RGB-based approaches. This contributes to improved robustness and reliability, even if it does not always result in the highest accuracy values~\cite{24,21}.

\subsection{ROC Curve Analysis}
\begin{figure}[H]
\centering

\begin{subfigure}{0.3\textwidth}
\includegraphics[width=\linewidth]{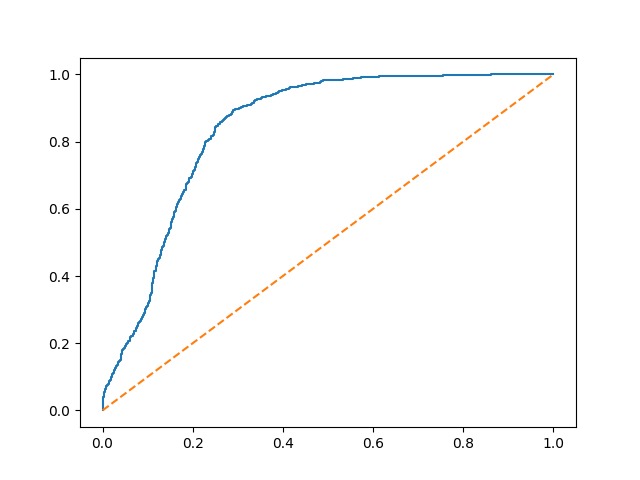}
\caption{DenseNet121}
\end{subfigure}
\hfill
\begin{subfigure}{0.3\textwidth}
\includegraphics[width=\linewidth]{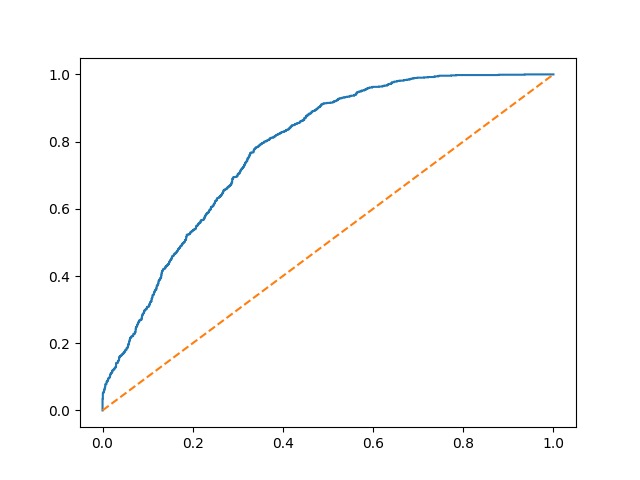}
\caption{VGG16}
\end{subfigure}
\hfill
\begin{subfigure}{0.3\textwidth}
\includegraphics[width=\linewidth]{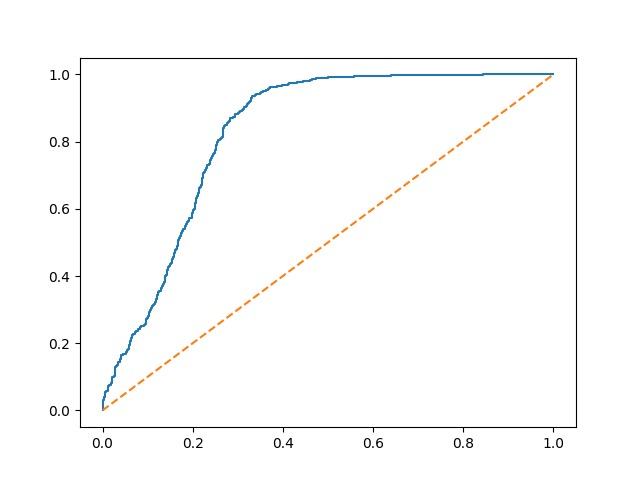}
\caption{ResNet50}
\end{subfigure}

\vspace{0.3cm}

\begin{subfigure}{0.3\textwidth}
\includegraphics[width=\linewidth]{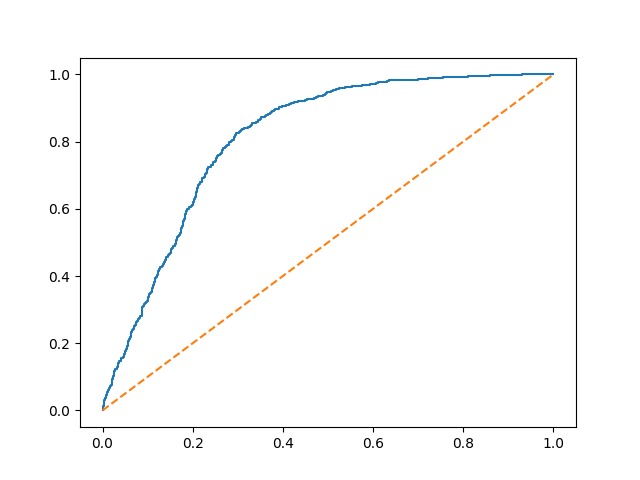}
\caption{EfficientNetB0}
\end{subfigure}
\hfill
\begin{subfigure}{0.3\textwidth}
\includegraphics[width=\linewidth]{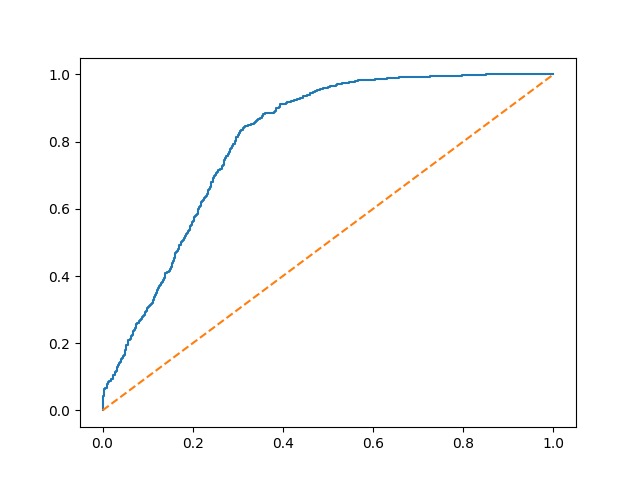}
\caption{MobileNet}
\end{subfigure}
\hfill
\begin{subfigure}{0.3\textwidth}
\includegraphics[width=\linewidth]{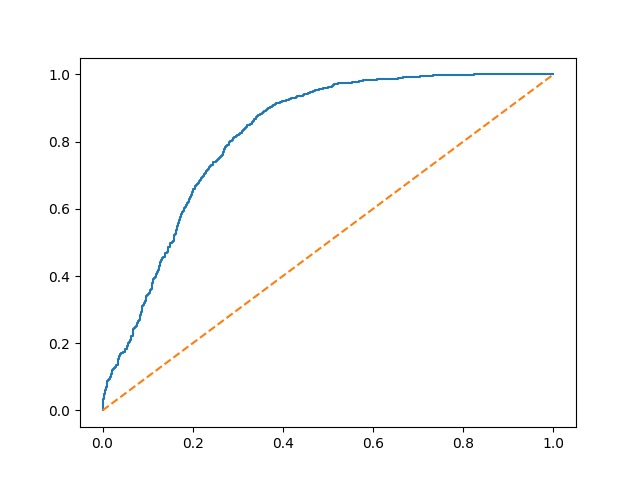}
\caption{InceptionV3}
\end{subfigure}

\caption{ROC curves illustrating the classification performance of each evaluated CNN model.}
\label{fig:roc_all}
\end{figure}

As shown in Fig.~\ref{fig:roc_all}, DenseNet121 and ResNet50 demonstrate superior ROC performance, characterized by higher true positive rates at lower false positive rates compared to other models.

The ROC curve illustrates the trade-off between true positive rate (TPR) and false positive rate (FPR)~\cite{1}. The AUC metric summarizes the overall discriminative capability of the model~\cite{5}.

The results indicate that DenseNet121 achieved the highest AUC, followed by ResNet50 and InceptionV3. These findings demonstrate that DenseNet121 and ResNet50 provide better generalization capability across varying thresholds~\cite{5,7}.

These results indicate that models with higher AUC and balanced performance tend to provide more reliable and stable classification behavior in real-world forensic applications~\cite{2,19}. Moreover, the ROC curves support the effectiveness of adaptive threshold selection, as models with higher AUC values provide more flexible operating points for determining optimal thresholds~\cite{30}. This enables better calibration of decision boundaries based on application-specific requirements, which is a well-established concept in classifier evaluation and decision theory~\cite{5,30}.


\subsection{Confusion Matrix Analysis}
\begin{figure}[H]
\centering

\begin{subfigure}{0.3\textwidth}
\includegraphics[width=\linewidth]{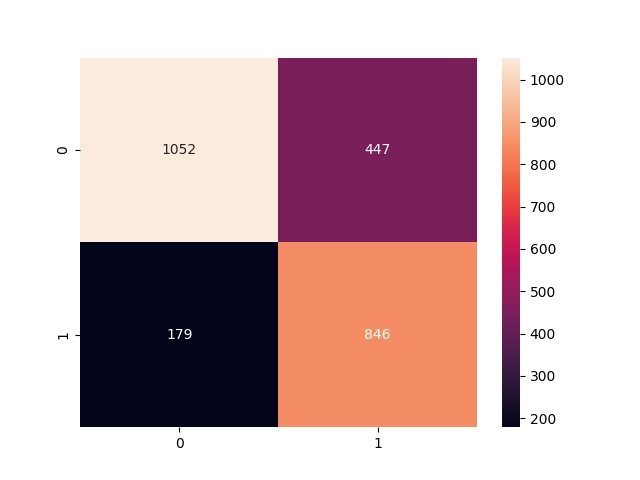}
\caption{DenseNet121}
\end{subfigure}
\hfill
\begin{subfigure}{0.3\textwidth}
\includegraphics[width=\linewidth]{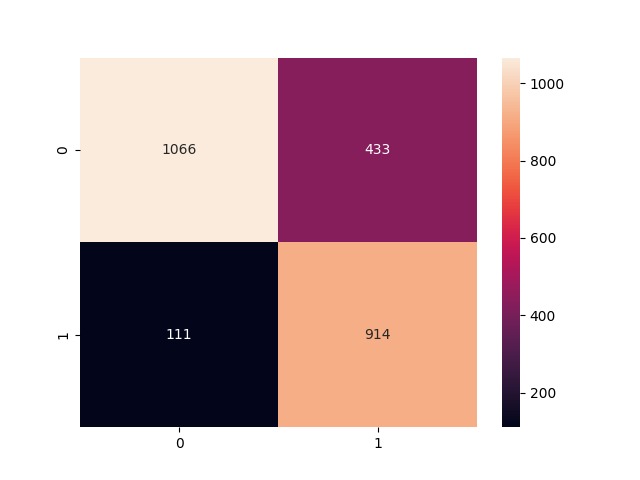}
\caption{VGG16}
\end{subfigure}
\hfill
\begin{subfigure}{0.3\textwidth}
\includegraphics[width=\linewidth]{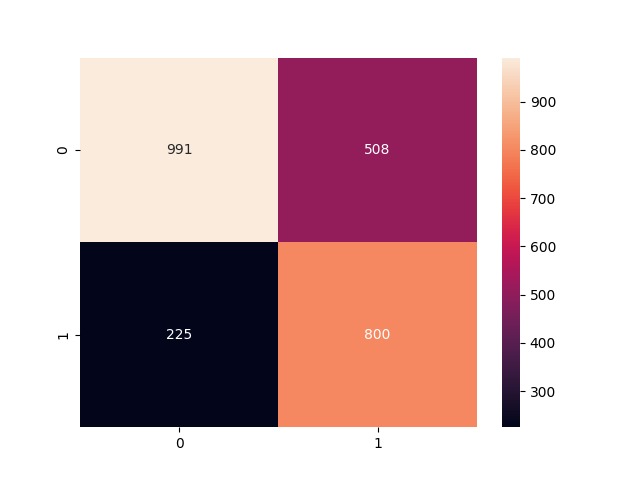}
\caption{ResNet50}
\end{subfigure}

\vspace{0.3cm}

\begin{subfigure}{0.3\textwidth}
\includegraphics[width=\linewidth]{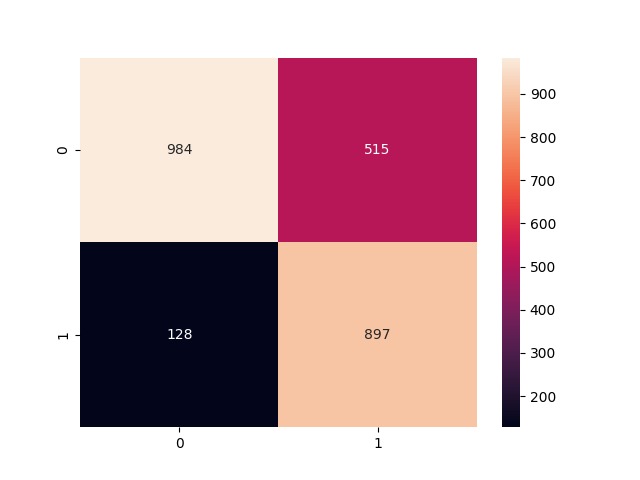}
\caption{EfficientNetB0}
\end{subfigure}
\hfill
\begin{subfigure}{0.3\textwidth}
\includegraphics[width=\linewidth]{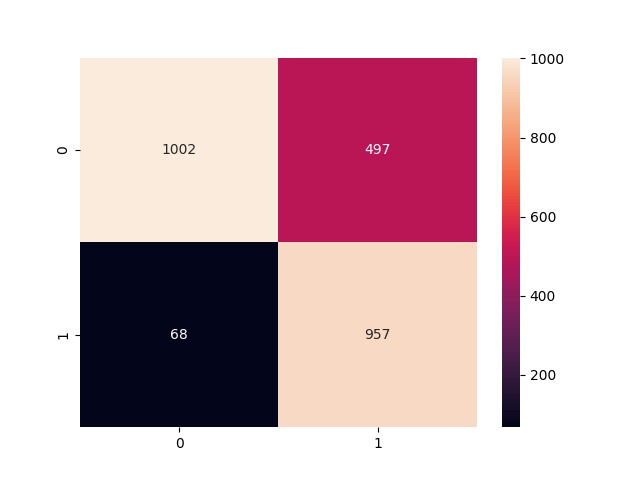}
\caption{MobileNet}
\end{subfigure}
\hfill
\begin{subfigure}{0.3\textwidth}
\includegraphics[width=\linewidth]{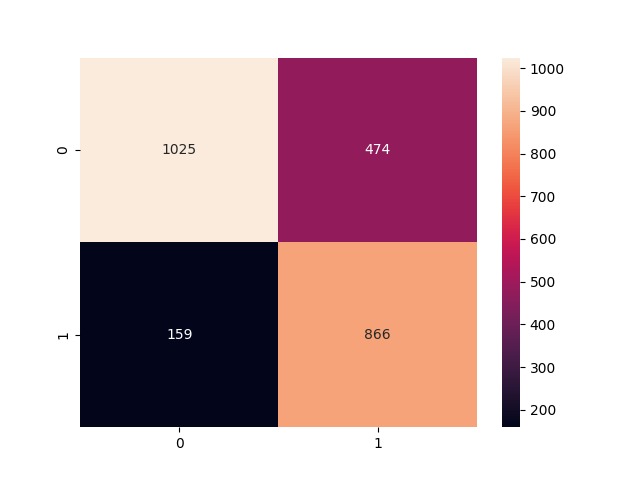}
\caption{InceptionV3}
\end{subfigure}

\caption{Confusion matrices for each evaluated CNN model.}
\label{fig:cm_all}
\end{figure}

As illustrated in Fig.~\ref{fig:cm_all}, the confusion matrices provide detailed insights into classification performance by explicitly illustrating the distribution of true positives, false positives, true negatives, and false negatives for each model. The confusion matrix results further highlight the impact of model-specific threshold selection on classification behavior. Models with lower optimal threshold values, such as ResNet50 (0.338), exhibit reduced false negative rates, as they shift the decision boundary towards the positive class, effectively reducing false negatives. This behavior is particularly desirable in digital forensic applications, where missing manipulated images constitutes a critical error in forensic analysis~\cite{2,19}.

In contrast, models with higher threshold values, such as EfficientNetB0 (0.461), demonstrate relatively higher false negative rates, as stronger confidence is required for positive classification. Such variations in classification behavior are consistent with prior studies emphasizing the role of decision threshold calibration in controlling the trade-off between sensitivity and specificity in machine learning systems~\cite{5,30}. These findings confirm that threshold selection directly influences the trade-off between false positives and false negatives, and reinforce the importance of adaptive, model-specific thresholding in improving classification reliability~\cite{7,21}.

\begin{enumerate}

\item \textbf{DenseNet121:}
\begin{itemize}
\item \textbf{Strengths:} Highest accuracy and strong detection capability.
\item \textbf{Weaknesses:} Moderate false negative rate.
\item \textbf{Insights:} Effective but may require optimization for critical detection cases.
\end{itemize}

\item \textbf{VGG16:}
\begin{itemize}
\item \textbf{Strengths:} Basic classification capability.
\item \textbf{Weaknesses:} High misclassification rates.
\item \textbf{Insights:} Limited ability to capture fine-grained artifacts.
\end{itemize}

\item \textbf{ResNet50:}
\begin{itemize}
\item \textbf{Strengths:} Most balanced performance with reduced false negatives.
\item \textbf{Weaknesses:} Slightly lower accuracy than DenseNet121.
\item \textbf{Insights:} Most suitable model for real-world forensic applications.
\end{itemize}

\item \textbf{EfficientNetB0:}
\begin{itemize}
\item \textbf{Strengths:} Moderate performance.
\item \textbf{Weaknesses:} Difficulty detecting subtle manipulations.
\item \textbf{Insights:} Requires further improvement.
\end{itemize}

\item \textbf{MobileNet:}
\begin{itemize}
\item \textbf{Strengths:} Efficient and lightweight.
\item \textbf{Weaknesses:} Lower discriminative capability.
\item \textbf{Insights:} Suitable for resource-constrained environments.
\end{itemize}

\item \textbf{InceptionV3:}
\begin{itemize}
\item \textbf{Strengths:} Competitive performance.
\item \textbf{Weaknesses:} Does not outperform DenseNet121 or ResNet50.
\item \textbf{Insights:} Larger input size does not guarantee better performance.
\end{itemize}

\end{enumerate}
\section{Discussion}

This study aimed to explore the application of transfer learning techniques for improving the detection of digital image forgery. Based on the experimental results, the following conclusions can be drawn:

\begin{itemize}

\item RQ1 investigates how transfer learning techniques can be effectively applied to enhance the accuracy and scalability of digital image forgery detection. The results indicate that DenseNet121 achieved the highest overall accuracy (78.4\%) and AUC (0.841), indicating strong discriminative capability. In addition, ResNet50 achieved the highest MCC value (0.598), reflecting the most reliable and balanced classification performance among the evaluated models.

The results emphasize that transfer learning-based CNN architectures are effective in capturing manipulation artifacts, particularly when combined with feature enhancement strategies. Unlike earlier observations in the literature, InceptionV3 did not outperform other models in this study, despite its larger input size and multi-scale feature extraction capability.

The comparative performance of DenseNet121 and ResNet50 further confirms the potential of pretrained CNNs for forgery detection, consistent with recent studies that combine CNN architectures and hybrid deep learning models for detecting image manipulations~\cite{18,20}. However, the relatively lower performance of models such as VGG16 and EfficientNetB0 indicates that architectural design and feature sensitivity play a critical role in detecting subtle manipulation artifacts.

\end{itemize}

Furthermore, the integration of compression-based feature enhancement (FDIFF) contributes to improved visibility of manipulation artifacts, enabling models to better distinguish between authentic and forged images~\cite{25}. In addition, the use of adaptive threshold optimization based on the Youden Index improves classification reliability by reducing critical errors, particularly false negatives~\cite{30}.

\textbf{Performance Interpretation:}

Although the achieved accuracy is lower than some existing studies, this does not necessarily imply inferior performance. Many previous works primarily focus on maximizing accuracy, often overlooking other critical performance metrics~\cite{7,21}. In contrast, this study emphasizes a more comprehensive evaluation using MCC and AUC, which provide a more realistic assessment of classification performance~\cite{5}.

In digital image forensics, false negatives represent critical errors, as undetected manipulated images may lead to serious consequences such as misinformation and security risks~\cite{2,19}. Therefore, models that achieve more balanced performance, such as ResNet50, are more suitable for real-world applications despite not having the highest accuracy.

Furthermore, the proposed framework introduces a more challenging evaluation setting by incorporating compression-based feature enhancement and adaptive threshold optimization. This setup more accurately reflects real-world scenarios where manipulation artifacts are subtle and difficult to detect, which may lead to lower accuracy values compared to simplified experimental settings~\cite{24,21}.

\textbf{RQ1 Answer:}

The results clearly answer RQ1 by demonstrating that transfer learning combined with artifact-aware feature enhancement (FDIFF) and adaptive threshold optimization improves classification reliability, robustness, and the ability to detect subtle manipulation artifacts, particularly by reducing false negatives~\cite{7,25,30}.

Several studies have explored image forgery detection using deep learning. Table~\ref{tab:comparison} provides a comparative analysis of the present study with other related works:

\begin{table}[t]
\centering
\caption{Comparative Analysis with Other Related Works}
\label{tab:comparison}
\renewcommand{\arraystretch}{1.2}

\begin{tabular}{p{2.5cm} p{1.6cm} p{2.4cm} p{1.6cm} p{3cm}}
\hline
\textbf{Study} & \textbf{Dataset} & \textbf{Method} & \textbf{Accuracy (\%)} & \textbf{Remarks} \\
\hline

This Study & CASIA v2.0 & Transfer Learning (DenseNet121 / ResNet50) & 78.4 & Balanced and reliable performance with improved robustness. \\

Elaskily et al. (2021)~\cite{3} & Custom Dataset & ConvLSTM & 85.32 & Designed mainly for copy-move forgery detection. \\

Bappy et al. (2019)~\cite{12} & CASIA v1.0 & Hybrid CNN-LSTM & 87.90 & Detects and localizes manipulated regions. \\

Khalil et al. (2023)~\cite{16} & CASIA v2.0 & Transfer Learning (EfficientNet) & 88.52 & Demonstrates effectiveness of pretrained CNN models. \\

Anwar et al. (2023)~\cite{6} & Multiple Datasets & Deep Feature Extraction & 90.10 & Combines handcrafted and deep features. \\

Nirmalapriya et al. (2023)~\cite{18} & CASIA v2.0 & ASCA-SqueezeNet & 91.40 & Hybrid deep learning model for improved robustness. \\

\hline
\end{tabular}

\end{table}
\FloatBarrier

Although the reported accuracy in this study is lower than some existing works, this difference can be attributed to the more realistic and challenging evaluation setup adopted in this research. Unlike many previous studies that focus primarily on maximizing accuracy under controlled conditions, this study emphasizes balanced and reliable performance using multiple evaluation metrics.
Moreover, recent studies have shown that deep learning-based forensic models can benefit from artifact-aware feature learning and anomaly detection strategies~\cite{19,21,24}.

In particular, the incorporation of compression-based feature enhancement and adaptive threshold optimization introduces additional complexity to the classification task, making it more representative of real-world scenarios. Furthermore, the use of metrics such as MCC and AUC provides a more comprehensive evaluation, especially in cases where minimizing false negatives is critical. Therefore, the proposed approach prioritizes reliability and robustness over accuracy alone. The observed variation in optimal threshold values further supports the necessity of model-specific decision calibration, as different architectures exhibit distinct confidence distributions. This variation can be attributed to differences in feature representation capacity, network depth, and sensitivity to compression artifacts, which directly influence the probability outputs of each model. Consequently, architectures such as ResNet50, which produce more conservative probability estimates, benefit from lower threshold values, while others require higher thresholds to achieve optimal performance.

\textbf{FDIFF Contribution Analysis:}  

While a dedicated ablation study comparing raw RGB inputs and the proposed FDIFF representation is not included in this work, the effectiveness of the FDIFF-based approach is supported by its ability to emphasize compression-related artifacts, which are widely recognized as key indicators in digital image forgery detection~\cite{25}. By explicitly enhancing recompression inconsistencies, the proposed representation facilitates the learning of more discriminative features by CNN models~\cite{21,24}. 

Future work will focus on conducting a comprehensive ablation analysis to quantitatively assess the individual contribution of the FDIFF representation and further validate its effectiveness.

\section{Conclusion}

This study provides a comprehensive evaluation of transfer learning–based approaches for digital image forgery detection, highlighting their effectiveness in identifying subtle manipulation artifacts. By leveraging pretrained convolutional neural network (CNN) architectures, including DenseNet121, VGG16, ResNet50, EfficientNetB0, MobileNet, and InceptionV3, the proposed framework effectively captures fine-grained differences between authentic and manipulated image regions, even under challenging post-processing operations such as rotation, scaling, and compression.

The experimental findings indicate that integrating compression-based feature representation with transfer learning substantially improves the detection of manipulation artifacts that are often difficult to capture using conventional approaches. Among the evaluated models, DenseNet121 achieved the highest overall accuracy (78.4\%) and AUC (0.841), reflecting strong discriminative capability. In contrast, ResNet50 achieved the highest MCC value (0.598), indicating more balanced and reliable classification performance.

The results emphasize that accuracy alone is insufficient for evaluating model performance in digital forensics. In particular, minimizing false negatives is of critical importance, as undetected manipulated images may lead to significant consequences, including misinformation and security vulnerabilities. Therefore, models that provide more balanced performance, such as ResNet50, are better suited for real-world forensic applications.

Furthermore, the analysis reveals that increased architectural complexity or larger input sizes, as observed in InceptionV3, do not necessarily translate into improved performance. Instead, the ability to effectively represent manipulation artifacts and adopt robust decision strategies plays a more decisive role in achieving reliable and robust detection.

A key contribution of this study lies in the integration of compression-based feature enhancement (FDIFF) with adaptive threshold optimization using the Youden Index. This combination improves classification reliability by enhancing the balance between sensitivity and specificity while reducing critical errors. In addition, the use of model-specific adaptive thresholds highlights the importance of decision boundary calibration, demonstrating that threshold optimization is as crucial as feature extraction in deep learning–based forgery detection systems.

Despite the promising results, several limitations should be acknowledged and addressed in future work. Certain models, such as VGG16 and EfficientNetB0, exhibited reduced sensitivity to subtle manipulation patterns, suggesting the need for improved architectural design and feature representation. Moreover, while transfer learning reduces computational cost, its dependence on pretrained weights from general-purpose datasets may limit performance in domain-specific forensic applications. Another limitation of this study is the use of the test set as validation data during training, which may introduce a degree of bias in performance estimation.

Future work will focus on extending the proposed framework through ensemble learning strategies, hybrid architectures, and advanced data augmentation techniques to further enhance generalization performance. These directions are discussed in detail in Section VII.

\section{Future Work}

Although the proposed framework demonstrates promising performance in digital image forgery detection, several directions can be explored to further enhance its effectiveness and applicability.

First, future studies may include an ablation analysis to quantitatively evaluate the individual contributions of the FDIFF representation and the hybrid RGB–FDIFF input strategy. Such analysis would provide deeper insights into the impact of compression-based feature enhancement on model performance.

Second, the integration of explainable artificial intelligence (XAI) techniques, such as Grad-CAM, can be investigated to improve model interpretability and transparency, which are critical for forensic and legal applications.

Third, the proposed framework can be extended to more challenging datasets and real-world scenarios, including DeepFake detection and multi-manipulation cases. Evaluating the model on diverse datasets would improve its generalization capability.

In addition, ensemble learning strategies and hybrid architectures can be explored to combine the strengths of multiple deep learning models and further improve detection performance~\cite{18,20,24}.

Finally, future work may focus on developing lightweight and efficient models suitable for deployment in real-time and resource-constrained environments, such as mobile and edge-based forensic systems.

\section*{Acknowledgment}

This study is supported by the Marmara University Scientific Research Projects Commission (BAPKO) under Project ID: 12428. The authors would like to thank Marmara University for providing the necessary infrastructure and support for this research.

\end{document}